\pdfoutput=1

\documentclass[11pt]{article}

\usepackage{acl}

\usepackage{times}
\usepackage{latexsym}
\usepackage{adjustbox}
\usepackage[T1]{fontenc}

\usepackage[utf8]{inputenc}

\usepackage[activate={true,nocompatibility}]{microtype}

\usepackage{inconsolata}

\usepackage{graphicx}
\usepackage[smallerops]{newtxmath}
\usepackage{enumitem}
\usepackage{booktabs}
\usepackage{listings}
\usepackage{algorithm}
\usepackage{algorithmicx}
\usepackage{algpseudocode}
\usepackage{xspace}

\usepackage[disable,textsize=scriptsize,textwidth=2cm]{todonotes}

\DeclareMathAlphabet{\mathcal}{OMS}{cmsy}{m}{n}
\DeclareMathAlphabet{\mathbb}{U}{msb}{m}{n}
\setlist[itemize]{leftmargin=*,itemsep=0cm,topsep=0.2cm}

\newcommand{\calflow}{\textsc{SMCalFlow}}
\newcommand{\treedst}{\textsc{TreeDST}}
\newcommand{\mtop}{\textsc{MTOP}}

\newcommand{\Dset}{\mathcal{D}}
\newcommand{\fEnc}{\mathbf{F}_{\rm enc}}

\newcommand{\fQ}{\mathbf{Q}}
\newcommand{\rStep}{R_{\rm step}}
\newcommand{\rIter}{R_{\rm iter}}

\newcommand{\IterR}{\textsc{IterR}\xspace}

\usepackage{colortbl}
\definecolor{LightCyan}{rgb}{0.53,0.88,0.91}

\title{Learning to Retrieve Iteratively for In-Context Learning}

\author{
  Yunmo Chen\thanks{~Johns Hopkins University; performed while interning at Microsoft.}, Tongfei Chen, Harsh Jhamtani, Patrick Xia, Richard Shin\thanks{~Google; performed while at Microsoft.} \\
  \bf Jason Eisner, Benjamin Van Durme \\
  Microsoft \\
  {\small\texttt{yunmo@jhu.edu, \{tongfeichen,hjhamtani,patrickxia,jeisner,ben.vandurme\}@microsoft.com}}
}

\begin{document}
\maketitle
\begin{abstract}
We introduce \emph{iterative retrieval}, a novel framework that empowers retrievers to make iterative decisions through \emph{policy optimization}. Finding an optimal portfolio of retrieved items is a combinatorial optimization problem, generally considered NP-hard. This approach provides a learned approximation to such a solution, meeting specific task requirements under a given family of large language models (LLMs). We propose a training procedure based on reinforcement learning, incorporating feedback from LLMs. We instantiate an iterative retriever for composing in-context learning (ICL) exemplars and apply it to various semantic parsing tasks that demand synthesized programs as outputs. By adding only 4M additional parameters for state encoding, we convert an off-the-shelf dense retriever into a stateful iterative retriever, outperforming previous methods in selecting ICL exemplars on semantic parsing datasets such as \calflow, \treedst, and \mtop. Additionally, the trained iterative retriever generalizes across different inference LLMs beyond the one used during training.
\end{abstract}

\section{Introduction}
\label{sec:introduction}

A significant emergent capability of large language models (LLMs) is \emph{in-context learning} \citep[ICL;][]{BrownMRSKDNSSAA20}, which facilitates few-shot learning. In ICL, a set of \emph{exemplars}\footnote{~An exemplar is a tuple of input and output, demonstrating the mapping relationship between the two.} is usually provided to build the mapping relationship between inputs and outputs. These exemplars can either be hand-crafted and fixed or retrieved from a training set. However, if retrieving from the dataset, the retrievers used in such applications are typically off-the-shelf models (e.g., Contriever \citep{IzacardCHRBJG22}) that do not consider interactions among retrieved items when multiple targets are required, nor the specific characteristics of the inference LLMs and downstream task requirements. Research \citep[\emph{i.a.}]{gao-etal-2021-making,liu-etal-2022-makes,lu-etal-2022-fantastically} has shown that ICL is sensitive to both the exemplars provided and their order within prompts. Off-the-shelf retrievers, which generally rank items based solely on semantic similarity \citep[\textit{i.a.}]{lee-etal-2019-latent,ReimersG19}, do not ensure optimal conditions for either criterion, leading to suboptimal performance in downstream LLM generation. Hence, there is a need for a retriever capable of constructing a portfolio of items tailored to achieve optimal generation with LLMs.

\begin{figure}[t]
    \centering
    \includegraphics[width=\linewidth]{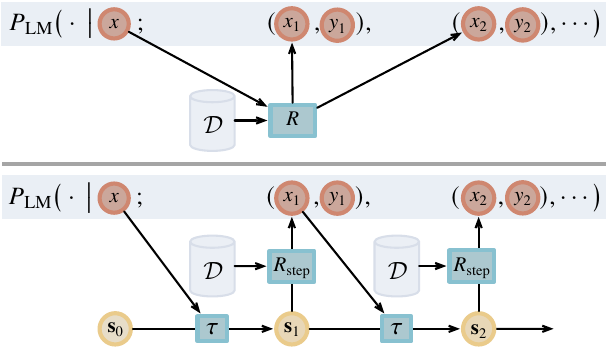}
    \caption{Above: ICL under a single retriever call. Below: ICL under our proposed iterative retriever. }
    \label{fig:iter-retriever}
    \vspace{-0.4cm}
\end{figure}

We propose \emph{iterative retrieval} to address this problem. Unlike traditional retrievers that perform a single call to obtain a list of similar items ordered by their similarities, iterative retrieval involves a sequence of retrieval calls, each using different query vectors. This makes the retriever stateful, maintaining an \emph{internal state}. The process can be likened to navigating the encoding space of exemplars, with each step adjusting direction based on previously selected exemplars, thus building a trajectory of exemplar selections.

This approach can be formulated as \emph{Markov decision processes} (MDPs). At each step, the \emph{action} taken by the retriever is a retrieval call that fetches (potentially multiple) documents from the dataset $\Dset$.\footnote{~The action space is at least as large as $\Dset$.} The policy is trained to optimally select exemplars at each step so that the overall trajectory maximizes the reward, leading to better ICL performance. By leveraging the LLMs as environments, we create simulators that allow a policy to roll out in the environment and receive feedback on the effectiveness of the composed prompts, measured by a reward (metric). Thus, exemplar selection and prompt composition can be framed as \emph{policy optimization} aimed at maximizing rewards, which can be addressed through reinforcement learning. 

We situate our study in in-context semantic parsing due to its difficulty, popularity, and practical value.\footnote{~Code generation is considered one of the most useful but challenging techniques in the era of LLMs. Some semantic parsing tasks share structural similarity with code generation and program synthesis.} 
We instantiate an iterative retriever and investigate the performance of policy learning under this setup. Our contributions include:
\begin{itemize}
    \item We propose a novel iterative retrieval framework that builds a portfolio of exemplars for ICL, considering both interactions among retrieved exemplars and their relationship with LLMs;
    \item We instantiate this iterative retriever for the in-context semantic parsing task and train its policy via reinforcement learning, demonstrating superior performance over strong baselines from prior work, thereby proving its effectiveness;
    \item Through a series of analyses, we provide insights into the behaviors of an iterative retriever initialized with an off-the-shelf retriever.
\end{itemize}

\section{Overview of an Iterative Retriever}
\label{sec:overview}

We consider the problem of \emph{in-context learning} (ICL): given a dataset $\Dset = \{(x_i, y_i)\}_i$ of \emph{exemplars}, a retriever $R$ retrieves a sequence of exemplars $R(x)$ based on input query $x$ and generate the answer $y$ based on the distribution $P_{\rm LM}(\cdot |x; R(x))$.

This retriever $R: \mathcal{X} \to \mathcal{D}^K$ retrieves an ordered list (of length $K$) of exemplars for the LM.  The goal of the retriever $R$ is to select a sequence of exemplars $((x_i, y_i))_{1 \le i \le K}$ such that the probability of the expected output $y$ is maximized:
\begin{equation}
  \label{eq:comb-opt-icl}
    \underset{\begin{subarray}{c}(x_i, y_i) \in \Dset\end{subarray}}{\arg\max} P_{\rm LM}(y |x; ((x_i, y_i))_{1 \le i \le K}).
\end{equation}

However, this is a \emph{combinatorial optimization} problem that is computationally infeasible to solve exactly. Much of prior work resort to selecting top-$k$ exemplars based on a scoring function $S$: 
\begin{equation}
    R(x) = \underset{(x^\prime, y^\prime) \in \Dset}{\arg\mathrm{top}_k} ~S(x, (x^\prime, y^\prime))
\end{equation}

Prior work has differed on the choice of the scoring function $S$: BM25 \cite{RoyTCSPED23}, coverage \cite{gupta-etal-2022-structurally}, etc.
However, such method did not model the interaction between the retrieved exemplars and the language model. We propose an iterative version, where we create a \emph{retrieval state} $s$, and for each step $i$ one exemplar $(x,y) \in \Dset$ is retrieved. This is an approximation to the optimization problem in \autoref{eq:comb-opt-icl}.
\begin{align}
    (x_i, y_i) &\leftarrow \rStep(s_i) \\
    s_{i+1} &\leftarrow \tau(s_i, (x_i, y_i))
\end{align}
After $K$ steps, the retrieved sequence would be $
    \rIter(x) = ((x_i, y_i))_{1 \le i \le K}$.
This formulation of an \emph{iterative retriever} naturally fits in the definition of a Markov decision process (MDP). Here, our decision process comprises of $(\Dset^*, \Dset, \tau, r)$, where
\begin{itemize}
    \item The state set $\Dset^*$ contains  exemplar sequences whose elements are in $\Dset$;
    \item The action set is just $\Dset$: each action selects one exemplar from $\Dset$. In theory, more than 1 exemplar can be selected at each step, but we proceed with just 1 exemplar for simplicity;
    \item The transition function $\tau: \Dset^* \times \Dset \to \Dset^*$ appends an exemplar to the existing sequence;
    \item The reward function $r: \Dset^* \times \Dset \to \mathbb{R}$ funnels signal from the LLM back to the retriever. It will be discussed in \S\ref{ss:reward}.
\end{itemize}

By situating our proposed iterative retriever under this RL scenario, we can utilize all sorts of RL techniques to train this retriever from the environment, which is the LLM itself.
In the next section, we instantiate a neural iterative retriever and situate it under a common task, namely \emph{semantic parsing}, under this ICL framework.

\begin{figure*}[t]
    \centering
    \includegraphics[width=\textwidth]{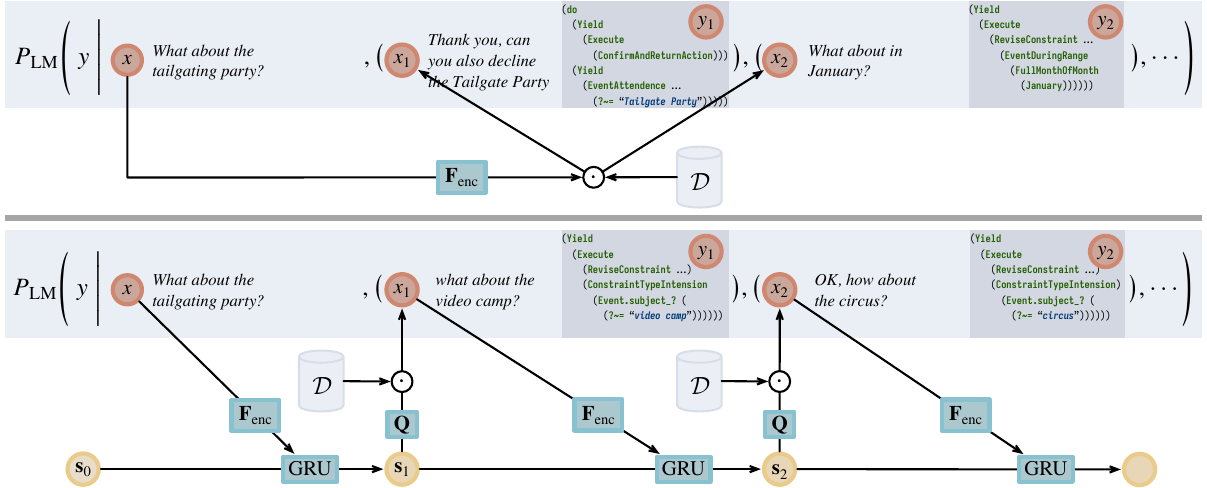}
    \caption{ICL prompt construction for an example in \calflow. \emph{Above:} ICL with BM25 as the retriever. \emph{Below:} An instance of our iterative retriever. BM25 retrieves examples that overlaps lexically with the query, whereas the trained iterative retriever is better at retrieving structurally similar exemplars since it is trained to maximize the probability of the LM generating the reference parse.}
    \label{fig:mdp}
\end{figure*}

\section{Instantiating an Iterative Retriever}
\label{sec:iter-retriever}

\begin{figure}[t]
    \centering
    \includegraphics[width=\linewidth]{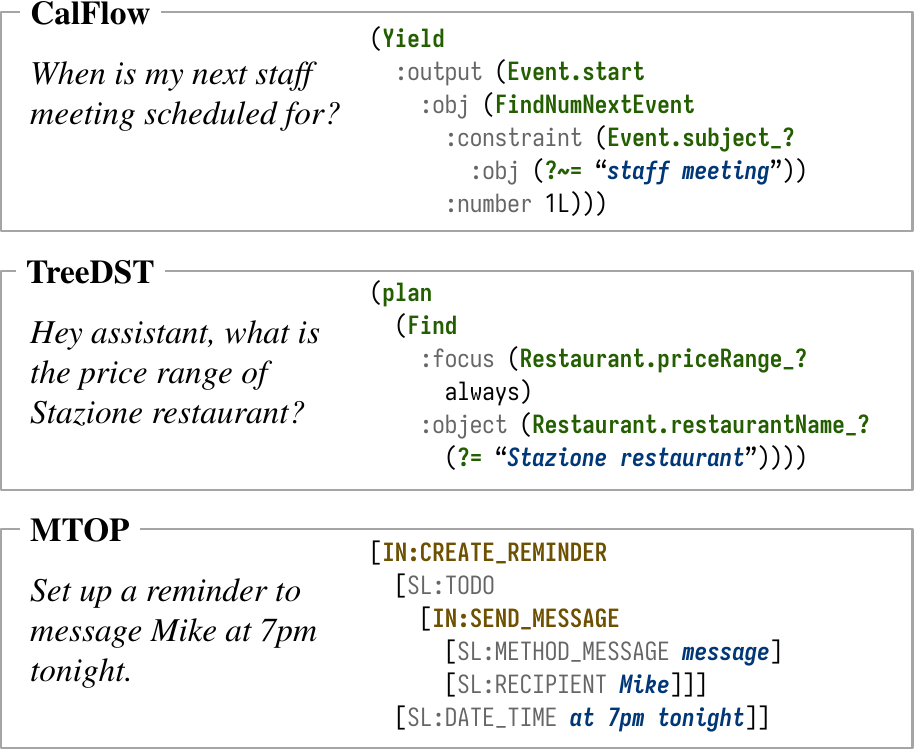}
    \caption{Samples of $(x, y)$ pairs for semantic parsing under different datasets used in this paper.}
    \label{fig:examples}
\end{figure}

We consider an \emph{instance} of in-context learning, namely few-shot semantic parsing. Given a natural language query $x$, a model is expected to output a semantic representation $y$ of $x$ given a sequence of exemplars (see \autoref{fig:examples}).

We instantiate a \emph{neural iterative retriever} based on the formulation we proposed above:
\begin{itemize}
    \item The state of the MDP, i.e. the sequence of exemplars, is modeled by a fixed-length vector $\mathbf{s} \in \mathbb{R}^d$. The initial state $\mathbf{s}_0$ is a parameter. 
    \item At each step 1 exemplar is retrieved. We define a \emph{policy distribution} that picks one exemplar from the training set $\Dset$, similar to \citet{Lu0CWZRCK23}:
    \begin{equation}
      \label{eq:policy}
        \pi((x_i, y_i) | s_i) \propto \exp(\fQ(\mathbf{s}_i) \cdot \fEnc(x_i) / \beta)
    \end{equation}
    where $\fQ:\mathbb{R}^d \to \mathbb{R}^d$ maps a state vector $\mathbf{s}_i$ to  a query vector $\mathbf{q}_i$,  $\fEnc: V^* \to \mathbb{R}^d$ is a text embedder that maps a text sequence into a vector, and $\beta$ is a temperature hyperparameter. In our experiments, $\fEnc$ is initialized with the weights of Contriever \cite{IzacardCHRBJG22}, a general-purpose text embedder trained for retrieval. 
    
    Under this policy, if we take greedy decoding, the retrieval step would just be
      \begin{align}
          \label{eq:r-step}
          (x_i, y_i) &\leftarrow \rStep(\mathbf{s}_i) = \underset{(x^\prime, y^\prime) \in \Dset}{\arg\max}~ \pi((x_i, y_i) | s_i) \nonumber \\
          &= \underset{(x^\prime, y^\prime) \in \Dset}{\arg\max}~ \fQ(\mathbf{s}_i) \cdot \fEnc(x_i).
      \end{align}
      This is a \emph{maximum inner product search} (MIPS) problem, and thus can be solved with a vector index such as FAISS \citep{douze2024faiss}.
    \item State transition is modeled by a gated recurrent unit \citep[GRU;][]{ChungGCB14} update:
      \begin{equation}
          \label{eq:transition}
          \mathbf{s}_{i+1} \leftarrow \mathrm{GRU}(\mathbf{s}_{i}, \fEnc(x_i))
      \end{equation}
      where the encoded vector of the retrieved exemplar $x_i$ is passed to the GRU to update the state.\footnote{~Using a Transformer decoder here results in more unstable training as we discovered in our experiments. See \S\ref{ss:ablation}.} 
\end{itemize}

Note that the only additional parameters we included in this neural iterative retriever is the state transition model, where we instantiate as a GRU.

This is different from a regular retriever, where a single retrieval call to the training set $R(x) =\arg\max_{(x,y) \in \Dset} \mathbf{q} \cdot \mathbf{F}_{\rm enc}(x)$ is made. The iterative retriever navigates the encoding space of exemplars, adjusting the query vector $\mathbf{q}^\prime$ at each step based on previously selected exemplars $s$, thus steering the search process to find new candidates. \autoref{fig:mdp} demonstrates the process of such an iterative retriever.
This stateful design allows for optimized retrieval results through iterative interactions, incorporating signals from both external sources (LLMs) and internal states (previously retrieved items tracked via state transitions).

\section{Training}
\label{sec:train-inference}

\paragraph{Environment Simulator}

To construct feedback (or reward) from the underlying LLMs, we treat LLMs as environments where actions are performed and evaluated. We design an iterative prompting schedule within this LLM environment to simulate the process of iterative retrieval and corresponding ICL prompt execution. At each step $i$, the current sequence of chosen exemplars, $s_i$, is turned into an LLM prompt using a predefined template,\footnote{~Refer to Appendix \ref{ss:prompt-template} for the template used in this work.}  then used for LLM generation. This schedule effectively simulates the real-world scenario of prompting LLMs, allowing us to observe various execution dynamics, such as generated hypotheses and their probabilities.

\paragraph{Reward Design}
\label{ss:reward}

Technically, if the final task metric were available, it can be used directly as the reward to optimize for. However, such a reward is often too coarse to reflect differences in partially correct results. For example, if the metric is exact match, which is common in semantic parsing tasks, the reward would simply be the Kronecker delta $\delta(y^*, y)$, yielding 1 only if the prediction $y$ exactly matches the reference $y^*$, and 0 otherwise.

Given that the LLM simulator provides access to the probabilities of generated sequences,\footnote{~This is generally accessible in many LLM inference implementations such as vLLM \cite{KwonLZ0ZY0ZS23}. For OpenAI-style APIs, this can be accessed using the ``\texttt{echo}'' parameter.} we employ a more general reward design that is not task-specific. Our reward leverages the LM completion probability of the reference sequence $P_{\rm LM}(y^*|x)$ \cite{shin-etal-2021-constrained,abs-2301-12652}, which captures subtle changes in the likelihood of the LM generating the target sequence with respect to changes in the input $x$. In ICL, more exemplars typically result in better performance before reaching saturation. Inspired by \citet{zhang-etal-2022-active}, We further refine the reward to reflect the \emph{increase in the likelihood} of the reference $y^*$ given the prompt. This is a proxy value that measure \emph{how much this exemplar contribute} to generating the reference parse. This design encourages the model to select exemplars that most significantly contribute to the final result given the existing exemplar sequence $s_i$:
\begin{align}
    r(s_i, x_i) &= P_{\rm LM} (y^* \mid x; s_i, (x_i, y_i)) \nonumber \\
    &- P_{\rm LM}(y^* \mid x; s_i).
\end{align}

\paragraph{Policy Optimization}

We employ \emph{proximal policy optimization} \citep[PPO;][]{SchulmanWDRK17} to train an iterative retriever for its stability and efficiency.\footnote{~We experimented with various other RL algorithms (including policy gradient~\citep{SuttonMSM99} and advantage actor critic \citep[A2C;][]{MnihBMGLHSK16}) and found that PPO is the most stable one for our scenario.} 
One core idea of PPO is to define a clipping term that controls the policy optimization process, so that variance is reduced. Given a trajectory $(x_1, \cdots, x_T)$, we have
\begin{align}\hspace*{-0.3cm}
        \mathcal{L}_i^{\text{clip}}(\theta) =\hat{\mathbb{E}}_i \big[ \min ( \rho_i , \mathrm{clip}_{\varepsilon}(\rho_i) ) \cdot \hat{A}_i ) \big],
\end{align}
where $\rho_i = \frac{\pi_\theta(x_i | s_i)}{\pi_{\theta_{\rm old}}(x_i | s_i)}$ is a probability ratio between action $x_i$\footnote{~$x_i$ describes that the action of an iterative retriever is to retrieve an exemplar from a candidate set, hence $x_i \in \Dset$.} performed against the current policy $\pi_\theta$ and old policy $\pi_{\theta_{\rm old}}$ at state $s_i$, $\mathrm{clip}_\varepsilon(\rho)$ clips $\rho$ to be within $(1-\varepsilon, 1+\varepsilon)$ and $\hat{A}$ is the advantage. 

Advantage $\hat{A}_i$ at step $i$ describes how much better it is to take a specific action $x_i$ at state $s_i$, over randomly selecting an action according to $\pi(x_i|s_i)$. To compute it, besides the neural model defined in \S\ref{sec:iter-retriever}, we follow common practice in reinforcement from human feedback \citep[RLHF;][]{abs-2403-17031} to add a single linear layer to serve as a state-value function $V(s) = \mathbf{v}\cdot\mathbf{s}$ that maps states to values. Generalized advantage estimate \citep[GAE;][]{SchulmanMLJA15} is then used for variance-reduced advantage estimation atop the learned state-value function:
\begin{align}
    \hat{A}_i &= \delta_i + (\gamma\lambda)\delta_{i+1} + \cdots + (\gamma\lambda)^{T-i+1}\delta_{T-1} \\
    \delta_i &= r_i + \gamma V(s_{i+1}) - V(s_i)
\end{align}
 where $r_i$ is the reward obtained at step $i$, $\gamma$ is the discount factor, $\lambda$ downweighs rewards corresponding to delayed effects. Following \citet{SchulmanWDRK17} on PPO in Actor-Critic style, we minimize the value function error term by a squared-error loss, with an additional entropy bonus term $-H$:
\begin{equation}
    L_i^{\rm PPO} =\mathbb{E}_i \big[\mathcal{L}_i^{\text{clip}}(\theta)+c_1\hat A_i^2(\theta) - c_2 H_{\pi_\theta}(s_t)\big]
\end{equation}
where $c_1$, $c_2$ are coefficients.

\paragraph{Sampling \& Collecting Experience}
\label{subsec:sampling-exps}

In a single retrieval step, the retriever selects an exemplar from a candidate set, with the policy $\pi_\theta(x_i | s_i)$ defining a probability distribution over candidates $x_i \in \Dset$. In this RL simulation, it is crucial to sample different actions at each step to enable the model to explore various trajectories and benefit from those that yield higher rewards. However, constructing the entire distribution and sampling from it at each step is computationally infeasible, especially when the number of candidates exceeds 100K. Furthermore, these distributions often exhibit a long-tailed nature, where many candidates have low scores, suggesting that a significant portion of candidates may be less similar and potentially less useful for ICL.

\begin{figure}
    \centering
    \includegraphics[width=0.8\linewidth]{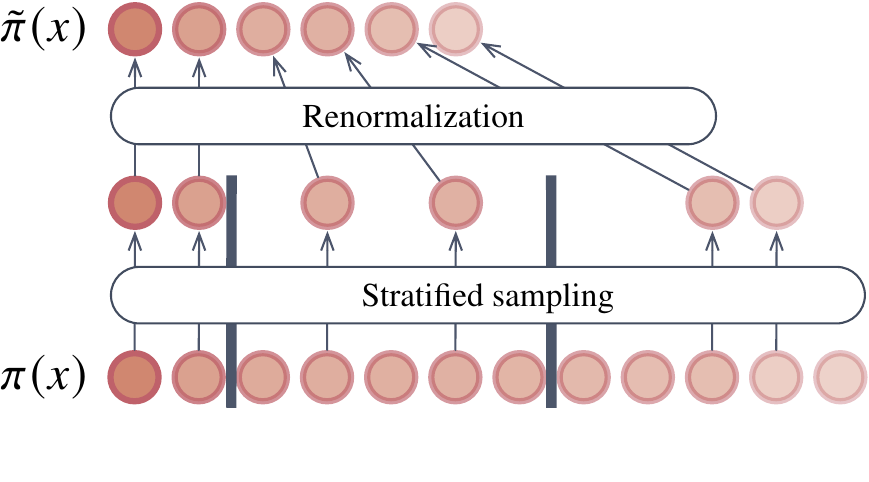}
    \vspace{-0.4cm}
    \caption{Stratified sampling employed in our approach. Our sampling method retains the top $k/N_s$ samples and split the rest into $(N_s - 1)$ strata to perform stratified sampling. The resulting $k$ samples are renormalized to construct action distribution.}
    \label{fig:stratfied-sampling}
    \vspace{-1em}
\end{figure}

To address these challenges and reduce the computational cost of sampling trajectories while managing the trade-offs between exploration and exploitation, we propose a stratified sampling ($N_s$ strata) method to construct a modified policy $\tilde\pi$ that contains $k$ candidates.

To start, we construct a buffer with top-$B$ exemplars retrieved with \autoref{eq:policy}. Retain the top $k/N_s$ samples in the policy. Split the rest into $(N_s - 1)$ strata, and sample $k/N_s$ from each. Combine all these selected exemplars and renormalize these scores with softmax (with temperature $\beta_{\rm renorm}$).
This method  enables the model to focus on more promising candidates while still allowing for exploration (see \autoref{fig:stratfied-sampling} for an illustration). 

During training, experience replay~\citep{Lin92} is employed to improve training efficiency. To collect experience, we run inference with the current policy fixed  on several training examples to generate trajectories. At each step, information such as policy, reward, and value is recorded. These trajectories are stored in a replay buffer, then shuffled and split into mini-batches for policy optimization. This approach allows the same experiences to be replayed multiple times,  reducing the number of required simulation runs.

\section{Experimental Setup}

\paragraph{Datasets}
We validate our pilot iterative retriever for ICL on a set of semantic parsing datasets, namely \calflow\ \citep{andreas-etal-2020-task}, \treedst\ \citep{cheng-etal-2020-conversational}, and \mtop~\citep[English portion only; ][]{li-etal-2021-mtop}, following the BenchClamp benchmark \cite{RoyTCSPED23}. Samples of representations are shown in \autoref{fig:examples}. For statistics, see Appendix \ref{ss:dataset-stats}.

\paragraph{Baselines}
We compare our iterative retriever (henceforth denoted as \textbf{\IterR}) with a range of off-the-shelf retrievers, including BM25 \citep{RobertsonZ09} and a dense encoder, Contriever \citep{IzacardCHRBJG22}. Additionally, we benchmark against two strong baselines from prior work on improving exemplar selection: \textbf{EPR}~\citep{rubin-etal-2022-learning} and \textbf{CEIL}~\citep{Ye0F0K23}. \textbf{EPR} is an efficient exemplar retrieval method for in-context learning (ICL) that leverages a scoring LM to label positive and negative training examples, then using this dataset for a contrastively learned dense retriever. \textbf{CEIL} uses \emph{determinantal point processes} (DPPs) to model the interaction between the given input and in-context exemplars.

For the EPR baseline, we replace the base dense retrieval encoder with Contriever instead of S-BERT~\citep{reimers-gurevych-2019-sentence} for fair comparison. Following \citet{Ye0F0K23}, we use the trained EPR model as initialization for CEIL. Similarly, the same EPR checkpoint is used to initialize the text encoder in \IterR. Note that in \IterR, we freeze the weights of the EPR encoder and \textit{only train the GRU-based state transition function, policy network, and value network}, resulting in 4M more parameters compared to the original Contriever ($110{\rm M} \to 114{\rm M}$).

For retrievers without iterative capabilities, we adapt them by taking only the top-$k$ retrieved items and keeping their original ranks. For EPR, CEIL, and \IterR, we selected the best performing model checkpoints on the validation set. All generation is run with \textbf{10} exemplars; i.e. $k=10$.

\paragraph{Generation with LLMs} 
The inference LLM is essential for executing input prompts to generate responses. In our experiments, we use \texttt{Llama-2-7b} to build the environment simulator and train the policy using its signals. With the learned policy, we investigate both intra-family and inter-family generalization by replacing the inference LLMs. For models within the same Llama-2 family, we explore various model sizes and finetuned versions, including \texttt{CodeLlama-70b-Instruct}~\citep{CodeLlama}, a model further fine-tuned for code generation. For inter-family experiments, we choose \texttt{Mistral-7b} \citep{Mistral7b}. For decoding configurations, we consistently use beam search with beam size 3 and sampling temperature 0.5.

\paragraph{Hyperparameters} Please refer to Appendix \ref{ss:hyperparam}.

\paragraph{Evaluation Metrics}
We follow prior work in evaluating semantic parsing \citep{RoyTCSPED23}, where  \emph{exact match} at $k$ (EM@$k$) is used. Exact match results for top-$k$ decoded hypotheses reflects beam search decoding used in LLMs, where multiple parsing results are generated simultaneously.

However, EM is a stringent metric, penalizing even minor mismatches. For instance, a parse with a substructure reordered differently (\texttt{a \&\& b}) from the reference (\texttt{b \&\& a}) is still correct but would score zero under EM. This is problematic in semantic parsing, where target parses are compositional, making it important to assess the correctness of substructures. Since \calflow\ and \treedst\ involve  deeply nested structures, we also adopt \emph{SMatch} \citep{cai-knight-2013-smatch}, following \citet{chen-etal-2023-unified}, to evaluate performance on  substructures.
SMatch is designed to evaluate AMRs \cite{langkilde-knight-1998-generation-exploits}. Generated code can be transformed to AMRs by treating each function's return value as an \emph{entity} and each argument to a function as a \emph{value}, where the parameter name is the \emph{relation}. See \autoref{app:smatch-eval} for details.

\section{Results \& Analyses}
\label{exp:main-results}

We evaluate the performance of different retrievers by comparing their downstream ICL performance on semantic parsing (\autoref{tab:main-results}).  \IterR outperforms all baselines across three datasets on all metrics.

The gain in EM is intuitive since it aligns with the training objective, which involves  the probability of generating target parses. The improvement in SMatch indicates that \IterR~optimizes retrieval results to improve compositionality to some extent, even with a simple objective.\footnote{~While a more dedicated reward design, such as incorporating various linearizations of target structures, might further enhance \IterR's performance. This work focuses on demonstrating the framework's effectiveness rather than dedicatedly optimizing for a specific task design.}

\begin{table}[t]
    \centering
    \small
    \begin{tabular}{l|ccc|ccc}
        \toprule
        \bf Retriever & \multicolumn{3}{c}{\bf Exact Match} & \multicolumn{3}{c}{\bf SMatch} \\
        \cmidrule(lr){2-4} \cmidrule(lr){5-7}
                 & @1 & @2 & @3 & P & R & F \\
        \midrule
        \multicolumn{7}{l}{\textbf{\calflow}} \\
        BM25 & 39.8 & 43.7 & 44.0 & 66.6 & 64.2 & 65.3 \\
        Contriever & 44.0 & 48.5 & 48.9 & 68.4 & 66.8 & 67.6 \\
        EPR & 48.5 & 52.0 & 52.3 & 73.3 & 76.7 & 75.0 \\
        CEIL & 51.1 & 54.2 & 55.8 & 74.9 & 75.2 & 75.1 \\
      \rowcolor{LightCyan}
        \textsc{IterR} & \bf 54.1 & \bf 58.4 & \bf 58.5 & \bf 76.6 & \bf 78.1 & \bf 77.3 \\
        \midrule
        \multicolumn{7}{l}{\textbf{\treedst}} \\
        BM25 & 50.8 & 56.1 & 56.6 & 81.8 & 81.8 & 81.8 \\
        Contriever & 54.7 & 60.4 & 61.0 & 83.3 & 82.5 & 82.9 \\
        EPR & 54.0 & 58.2 & 58.8 & 84.7 & 83.4 & 84.0 \\
        CEIL & 56.2 & 58.3 & 61.6 & 81.3 & 84.4 & 84.9 \\
      \rowcolor{LightCyan}
        \textsc{IterR} & \bf 58.2 & \bf 63.4 & \bf 63.8 & \bf 85.5 & \bf 85.8 & \bf 85.7 \\
        \midrule
        \multicolumn{7}{l}{\textbf{\mtop}} \\
        BM25 & 57.4 & 63.2 & 63.9 & - & - & - \\
        Contriever & 59.3 & 64.2 & 64.7 & - & - & - \\
        EPR & 62.3 & 68.8 & 69.2 & - & - & - \\
        CEIL & 63.6 & 69.4 & 69.8 & - & - & - \\
      \rowcolor{LightCyan}

        \textsc{IterR} & \bf 63.9 & \bf 70.9 & \bf 71.0 & - & - & - \\
        \bottomrule
    \end{tabular}
    \caption{Comparison of our approach, \IterR against baselines. ``EM@$k$'' denotes exact match at top-$k$; ``P'', ``R'' and ``F'' denote precision, recall, and $\rm F_1$ score respectively. Experiment results are run with 10 exemplars in the prompt, averaged over 3 inference runs, and significance tests using paired $t$-test confirm that the improvements over Contriever, EPR, and CEIL are statistically significant ($p < 0.05$).}
    \vspace{-0.5em}
    \label{tab:main-results}
\end{table}

\begin{figure*}[t]
    \centering
    \includegraphics[width=\linewidth]{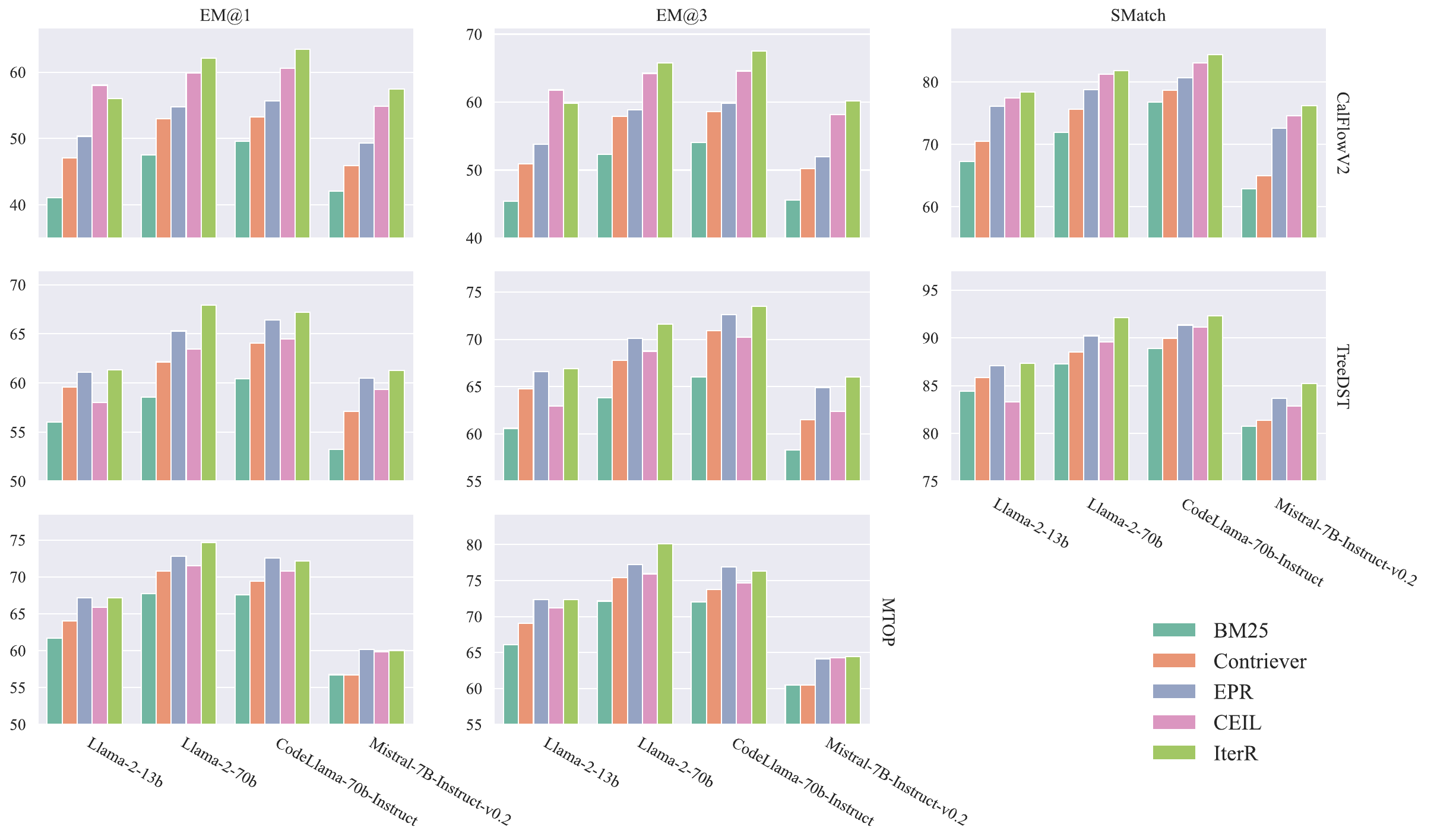}
    \caption{Performance comparisons on using various LLMs for inference (top row: \calflow; mid: \treedst; bottom: \mtop). Our \IterR used in these experiments are trained with \texttt{Llama-2-7b} but performs retrieval of ICL exemplars used on other LLMs.}
    \vspace{-1em}
    \label{fig:generalization-exps}
\end{figure*}

\paragraph{Generalization across Inference LLMs}

\IterR benefits from interactive training with an underlying LLM. While training incurs costs, these can be minimized by training only once, ideally using a smaller LM. Hence in this section we investigate the generalization capabilities of \IterR trained with a smaller LM $A$, but used for generation under a larger LM $B$. 

In the following experiments, \IterR is trained with \texttt{Llama-2-7b} as the environment, but used for \textbf{(a)} \emph{intra-family LMs}: variants within the \texttt{Llama-2} model family; and \textbf{(b)} \emph{inter-family LMs}:  Mistral \citep{Mistral7b} from a different model family. We follow the setups described in \S\ref{exp:main-results}, substituting only the LLM. As shown in \autoref{fig:generalization-exps}, \IterR significantly outperforms ($>1\%$ gain) baselines for $75\%$ of the settings and is comparable to a prior strong baseline (within $1\%$ in absolute performance) for $15\%$ of settings,    demonstrating its generalization within and beyond its own model family.

In intra-family generalization, performance metrics improve with larger model sizes, and \IterR consistently outperforms all baselines. This improvement is most evident with larger models such as \texttt{Llama-2-70b} and \texttt{CodeLlama-70b-Instruct}. For inter-family generalization, \IterR maintains its advantage across datasets, though this is less pronounced than within the same model family. This is expected, as the signal from LLM simulator is more representative for models sharing the same pre-training procedure. Notably, with Mistral, Contriever performs worse than BM25 on \mtop, but \IterR still shows improvement. This suggests that \IterR, comprising a frozen EPR and additional GRU layers, can learn task-specific abilities not present in the vanilla EPR.

\paragraph{ICL \& Number of Exemplars}

\begin{figure*}[t]
    \centering
    \includegraphics[width=\linewidth]{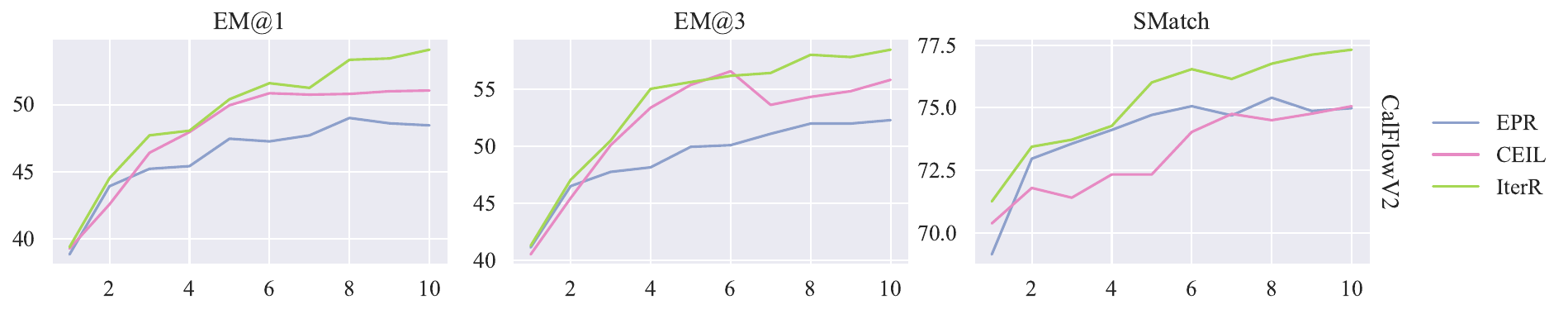}
    \vspace{-1em}
    \caption{Performance comparisons across the various numbers of exemplars used for ICL. }
    \vspace{-1em}
    \label{fig:num-exemplar-exps}
\end{figure*}

We investigated how the performance of \IterR changes with the number of exemplars ($\{1, \cdots, 10\}$) used for ICL on the \calflow\ dataset (\autoref{fig:num-exemplar-exps}). \IterR consistently outperforms baseline models across various metrics and numbers of exemplars, with one exception for the EM@3 metric when using 6 exemplars. This aligns with our training objective, where actions that boost performance at each step receive higher advantages. \IterR achieves comparable performance with fewer exemplars.

CEIL shows a similar trend in EM, but its SMatch performance lags significantly, indicating poorer quality in near-miss predictions compared to \IterR. Practically, this means our method allows for a trade-off between performance and cost, enabling effective ICL with fewer exemplars and reducing the number of tokens processed by LLMs.

\subsection{Ablation Study}
\label{ss:ablation}

\begin{table}[h]
    \centering \small
    \begin{tabular}{lcc}
    \toprule
        \bf Variant & \bf EM@1 & \bf SMatch-F \\
    \midrule
    Contriever & 44.0 & 67.6 \\
    \midrule
      \rowcolor{LightCyan}
        \textsc{IterR} & {\bf 54.1} & {\bf 77.3} \\
      \rowcolor{LightCyan}
        ~ \textit{$-$ EPR intialization} & 45.1 & 68.8 \\
      \rowcolor{LightCyan}
        ~ \textit{$-$ GRU; $+$ Transformer decoder} & 50.1 & 75.1 \\
              \rowcolor{LightCyan}
        ~ \textit{$-$ Stratified sampling} & 52.3 & 75.7 \\
    \bottomrule
    \end{tabular}
    \caption{Results on ablation study. \textit{$-$ EPR intialization} indicates the model is trained from Contriever instead of a EPR finetuned checkpoint. \textit{$+$ Transformer decoder} replaces GRU with a Transformer decoder. \textit{$-$ Stratified sampling} replaces the stratified sampling described in \autoref{fig:stratfied-sampling} with sampling directly from the buffer.}
    \vspace{-1em}
    \label{tab:ablation}
\end{table}
We further conduct ablation study on components of an iterative retriever, focusing on the \calflow\ dataset  and use \texttt{Llama-2-7b} while changing the configuration of the iterative retriever. Results are reported in \autoref{tab:ablation}.

\paragraph{EPR Initialization}
Although we follow prior work in using EPR as initialization for $\fEnc$, our iterative retriever is agnostic to the choice of base encoders for similarity search. 
Even without EPR initialization, our training procedure still improves performance against Contriever ($\approx 1\%$ gain under Contriever, but $\approx 6\%$ gain under EPR). We see that \IterR benefits more when using EPR initialization, significantly outperforming the baselines. We hypothesize that this advantage stems from two sources: (1) EPR is fine-tuned on the target dataset, making it more domain-specific; (2) EPR restructures the action space,  subsequently enhancing sample efficiency in RL training.

\paragraph{State Transition with Transformer Decoder}
In \S\ref{sec:iter-retriever}, we parameterize the state transition function in the iterative retriever with a GRU. To explore alternatives, we conducted an ablation experiment by replacing the GRU with a more powerful Transformer decoder, configured with 3 layers, 1024 hidden dimensions, with learnable positional encodings. Despite the increased expressiveness of the Transformer decoder, we observed a performance drop. During training, employing the warmup technique \cite{XiongYHZZXZLWL20} led to a trivial solution where the policy learned to predict a nearly fixed trajectory across test examples. Disabling the warmup stabilized the training but did not improve performance.
Developing a stabilized approach to train the Transformer decoder as a state encoder is beyond the scope of this work, as our focus is on demonstrating the overall framework of iterative retrieval rather than optimizing a specific model for the state transition function. Notably, even with the less powerful GRU, our iterative retriever successfully learns a policy that retrieves a more optimized sequence of ICL exemplars.

\paragraph{Effectiveness of Stratified Sampling}
To collect diverse experience from policy rollouts, we introduce a stratified sampling method (described in \S\ref{subsec:sampling-exps}) that balances the trade-off between exploration and exploitation.
We found that sampling from the raw policy in \autoref{eq:policy} results in a significant performance drop. Additionally, qualitative examination of several such distributions revealed a preference for exploitation over exploration, as similar items at the top of the retrieved list all had higher probabilities.

\section{Additional Related Work}

\paragraph{LLMs as Environment in RL}

\citet{Lu0CWZRCK23} used policy gradient to learn a dense retriever for ICL exemplar retrieval, but the state does not contain previously selected examples, and thus is not iterative and unable to model exemplar order.
\citet{zhang-etal-2022-active} used $Q$-learning RL for ICL exemplar reordering, with a similar reward design like ours. However, the proposed method does not extend to exemplar \emph{retrieval}, since the policy space is too large to be handled by $Q$-learning.

\paragraph{Few-shot Semantic Parsing}

Few-shot semantic parsing using LLMs has shown impressive capabilities in understanding new examples with minimal training data \citep{shin-etal-2021-constrained,shin-van-durme-2022-shot}. However, these parsers often struggle with generalization and fail to parse unobserved local structures due to their limited access to information encoded through exemplars \citep{bogin-etal-2022-unobserved}.
To this end, recent research has explored various approaches to improving exemplar selection. EPR \citep{rubin-etal-2022-learning}  used a proxy LM to score outputs from an unsupervised retriever, enabling better training of a dense retriever.   \citet{oren-etal-2021-finding}, \citet{gupta-etal-2022-structurally}, and \citet{levy-etal-2023-diverse} emphasize learning to select exemplars based on particular criteria, such as diversity measures and coverage of local structures, to enhance compositional generalization. 
While these approaches have shown performance improvements in semantic parsing tasks, these are highly based on heuristics constructed from researcher’s experience. Our approach could be seen as an \emph{automated} version (through RL) of seeking information useful for semantic parsing.

\section{Conclusion}
We proposed \emph{iterative retrievers} that iteratively builds a prompt to perform in-context learning. Such retrievers are framed as Markov decision processes and trained via policy optimization from LLM feedback, where the policy directs which exemplar to append to the existing exemplar sequence. Experiments on semantic parsing demonstrated performance gain of iterative retrievers over various datasets and state-of-the-art baselines, showing that they are able to construct prompts that improves in-context learning and downstream LLM generation.

\section*{Limitations}
In our instantiation of the iterative retriever, at each step a single exemplar is retrieved. One could envision multiple exemplars being retrieved at each step, thus making the RL trajectory shorter. This could make RL training easier and inference faster.

Our reward design depends on a particular linearization of the target structure. A more structured reward function may exhibit better training behavior and lead to better performance.

The encoder for queries in the iterative retriever is frozen in our current setup. A trainable query encoder that receives feedback from LLMs may be desired, but we left that for future work.

While we believe that semantic parsing / code generation is one of the most useful but challenging task for LLMs, as such is a representative task for ICL research, we have not tested the effectiveness of iterative retrievers under other LLM tasks.

\section*{Acknowledgements}

This work has been supported by the U.S. National Science Foundation under grant 2204926. Any opinions, findings, and conclusions or recommendations expressed in this article are those of the authors and do not necessarily reflect the views of the National Science Foundation.

\bibliography{anthology,custom}

\appendix

\newpage
\section{Experiment Details}

\subsection{Dataset Statistics}
\label{ss:dataset-stats}
\begin{table}[h]
    \centering\small
    \begin{tabular}{l|ccc}
    \toprule
        Dataset & Train & Dev & Test \\
    \midrule
        \calflow & 108,753 & 12,271 (500 used) & 13,496 \\
        \treedst & 121,652 & 22,910 (500 used) & 22,841 \\
        \mtop & 15,667 & 2,235 (500 used) & 4,386 \\
    \bottomrule
    \end{tabular}
    \caption{Dataset statistics.}
    \label{tab:data-stats}
\end{table}

\subsection{Hyperparameters}
\label{ss:hyperparam}
\begin{table}[h]
    \centering
    \adjustbox{max width=\linewidth}{
    \begin{tabular}{lc}
        \toprule
        Name & Search Bounds \\
        \midrule
        Encoder & $\{$ GTR-T5, \textbf{Contriever}, SBert $\}$ \\
        Learning rate &  $\{5\times 10^{-6}, 1\times 10^{-5}, {\bf 3\times 10^{-5}}, 5\times 10^{-5} \}$ \\
        LR scheduler & $\{$ \textbf{reduce-on-plateau}, cosine-annealing $\}$\\
        State transition & $\{$ \textbf{GRU}, LSTM, Transformer Decoder$ \}$\\
        $\beta_{\rm renorm}$ & $\{0.5, 1.0, {\bf 5.0}, 10.0\}$ \\
        $c_1$ & $\{0.1, 0.3, {\bf 0.5}, 0.7\}$ \\
        $c_2$ & $\{0, 0.005, {\bf 0.01}, 0.05, 0.1, 0.15\}$ \\
        \midrule 
        \midrule 
        $\gamma$ & 0.99 \\
        $\lambda$ & 0.95 \\
        Action buffer size & 768 \\
        PPO ratio cutoff & 1.2 \\
        PPO batch size & 128 \\
        Replay buffer size & 2048 \\
        Avg. training time & 24 hrs \\
        GPU used & 4 Nvidia V100 32 GB\\
        \# of parameters\textsuperscript{*} & $\sim$114M (w/ 110M frozen) \\
        \bottomrule
    \end{tabular}
    }
    \caption{Hyperparameters and other reproducibility information for \IterR. $\beta_{\rm renorm}$ is the temperature used to create a renormalized action distribution. $c_1$ and $c_2$ are coefficients used in the PPO loss. $\gamma$ and $\lambda$ are discount factors used in GAE.
    }
    \label{tab:hyperparams}
\end{table}

\subsection{Prompt Template}
\label{ss:prompt-template}
The prompt template used across all our experiments is shown in \autoref{tab:prompt-template}.
\begin{table}[H]
    \centering\small
    \begin{tabular}{l}
    \toprule
         Let's translate what a human user says\\
         into what a computer might say.  \\
         \\
         Human: $x_1$ \\
         Computer: $y_1$ \\
         \\
         $\cdots$ \\
         \\
         Human: $x_N$ \\
         Computer: $y_N$ \\
         \\
         Human: $x$ \\
         Computer: \\
    \bottomrule
    \end{tabular}
    \caption{Prompt template used in our experiments. This template will be instantiated as prompts when filled with retrieved exemplars $R(x) = ((x_1, y_1), \cdots, (x_N, y_N))$  and the test example $x$.}
    \label{tab:prompt-template}
\end{table}

\section{SMatch Evaluation}
\label{app:smatch-eval}

\lstset{basicstyle=\footnotesize\ttfamily,breaklines=true}
For evaluation of semantic parse or code generation on partial results, we utilize SMatch \cite{cai-knight-2013-smatch}. Generated code can be transformed to AMRs by treating each function's return value as an \emph{entity} and each argument to a function as a \emph{value}, where the parameter name is the \emph{relation}. An example is given below.

Consider the following parse in \calflow, expressed in Lisp:
\begin{lstlisting}[language=elisp]
  (Yield
    :output (Event.start
      :obj (FindNumNextEvent
        :constraint (Event.subject_?
          :obj (?~= "staff meeting"))
        :number 1L)))
\end{lstlisting}

This will be transformed into the following AMR:
\begin{figure}[H]
    \centering
    \includegraphics[width=0.25\textwidth]{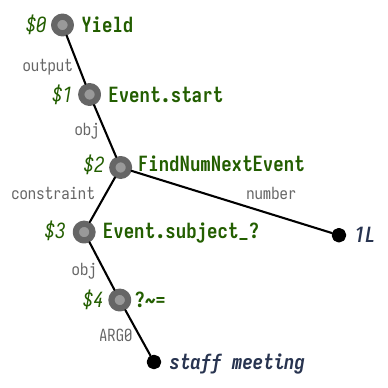}
    \caption{Example AMR based on the previous parse.}
    \label{fig:amr-example}
\end{figure}

This AMR can be easily converted to the following triples.
\begin{lstlisting}[language=Python]
  instance($0, Yield)
  output($0, $1)
  instance($1, Event.start)
  obj($1, $2)
  instance($2, FindNumNextEvent)
  constraint($2, $3)
  instance($3, Event.subject_?)
  obj($3, $4)
  instance($4, ?~=)
  ARG0($4, "staff meeting")
  number($2, 1L)
\end{lstlisting}

\end{document}